# Pivotal Pruning of Trade-offs in QPNs

**Silja Renooij, Linda C. van der Gaag,**
Department of Computer Science
Utrecht University
P.O. Box 80.089
3508 TB Utrecht, The Netherlands
{silja,linda}@cs.uu.nl

**Simon Parsons**
Department of Computer Science
University of Liverpool
Chadwick Building
Liverpool L69 7ZF, UK
s.d.parsons@csc.liv.ac.uk

**Shaw Green**
Department of Electronic Engineering
Queen Mary & Westfield College
University of London
London E1 4NS, UK
s.d.green@elec.qmw.ac.uk

## Abstract

Qualitative probabilistic networks have been designed for probabilistic reasoning in a qualitative way. Due to their coarse level of representation detail, qualitative probabilistic networks do not provide for resolving trade-offs and typically yield ambiguous results upon inference. We present an algorithm for computing more insightful results for unresolved trade-offs. The algorithm builds upon the idea of using pivots to zoom in on the trade-offs and identifying the information that would serve to resolve them.

## 1 INTRODUCTION

Qualitative probabilistic networks were introduced in the early 1990s for probabilistic reasoning with uncertainty in a qualitative way [Wellman, 1990]. A qualitative probabilistic network encodes variables and the probabilistic relationships between them in a directed acyclic graph. The encoded relationships basically represent influences on probability distributions. Each of these influences is summarised by a qualitative sign indicating the direction of shift in one variable's distribution occasioned by a shift in another variable's distribution. For probabilistic inference with qualitative networks, an elegant algorithm based upon the idea of propagating and combining signs is available [Druzdzel & Henrion, 1993a].

Qualitative probabilistic networks capture the relationships between their variables at a coarse level of representation detail. As a consequence, these networks do not provide for resolving trade-offs, that is, for establishing the net result of two or more conflicting influences on a variable's probability distribution. If trade-offs are represented in a qualitative network, then probabilistic inference will typically yield ambiguous results. Once an ambiguity arises, it will spread throughout most of the network upon inference, even if only a very small part of the network is truly ambiguous.

The issue of dealing with trade-offs in qualitative probabilistic networks has been addressed before by several researchers. S. Parsons has introduced, for example, the concept of categorical influences. A categorical influence is either an influence that serves to increase a probability to 1 or an influence that decreases a probability to 0, regardless of any other influences, and thereby resolves any trade-off in which it is involved [Parsons, 1995]. C.-L. Liu and M.P. Wellman have designed a method for resolving trade-offs based upon the idea of reverting to numerical probabilities whenever necessary [Liu & Wellman, 1998]. S. Renooij and L.C. van der Gaag have enhanced the basic formalism of qualitative probabilistic networks by distinguishing between strong and weak influences. Trade-off resolution during inference is then based on the idea that strong influences dominate over conflicting weak ones [Renooij & Van der Gaag, 1999]. These approaches to trade-off resolution are all based on a refinement of the representation used in the basic formalism.

In this paper, we present a new algorithm for dealing with trade-offs in qualitative probabilistic networks. Rather than resolving trade-offs by providing for a finer level of representation detail, our algorithm identifies the information that would serve to resolve the trade-offs present in a qualitative probabilistic network. From this information, a more insightful result than ambiguity is constructed.

Our algorithm for dealing with trade-offs builds upon the idea of zooming in on the part of a qualitative probabilistic network where the actual trade-offs reside. After a new observation has been entered into the network, probabilistic inference will provide the sign of the influence of this observation on the variable of interest, given previously entered observations. If this sign is ambiguous, then there are trade-offs present in the network. In fact, a trade-off must reside along the reasoning chains between the observation and the variable of interest. Our algorithm isolates these reasoning chains to constitute the part of the network that is relevant for addressing the trade-offs present. From this relevant part, an informative result is constructed for the variable of interest in terms of values for the variables in-volved and the relative strengths of the influences between them.

We believe that qualitative probabilistic networks can play an important role in the construction of quantitative probabilistic networks for real-life application domains, as well as for explanation of their reasoning processes. The construction of a probabilistic network typically sets out with the construction of the network's digraph. As the assessment of the various probabilities required is a far harder task, it is performed only when the network's digraph is considered robust. Now, by assessing signs for the influences modelled in the digraph, a qualitative network is obtained that can be exploited for studying the projected probabilistic network's reasoning behaviour prior to the assessment of its probabilities. For this purpose, algorithms are required that serve to derive as much information as possible from a qualitative probabilistic network. We look upon our algorithm as a first step to this end.

The paper is organised as follows. In Section 2, we provide some preliminaries concerning qualitative probabilistic networks. In Section 3, we introduce our algorithm for zooming in on trade-offs informally, by means of an example. The algorithm is discussed in further detail in Section 4. The paper ends with some concluding observations in Section 5.

## 2   PRELIMINARIES

A *qualitative probabilistic network* encodes the statistical variables from a domain of application and the probabilistic relationships between them in a directed acyclic graph $G = (V(G), A(G))$. Each node in the set $V(G)$ represents a statistical variable. Each arc in the set $A(G)$ can be looked upon as expressing a causal influence from the node at the tail of the arc on the node at the arc's head. More formally, the set of arcs captures probabilistic independence among the represented variables. We say that a chain between two nodes is blocked if it includes either an observed node with at least one outgoing arc or an unobserved node with two incoming arcs and no observed descendants. If all chains between two nodes are blocked, then these nodes are said to be *d-separated* and the corresponding variables are considered conditionally independent given the entered observations [Pearl, 1988].

A qualitative probabilistic network associates with its digraph $G$ a set $\Delta$ of qualitative influences and synergies [Wellman, 1990]. A *qualitative influence* between two nodes expresses how the values of one node influence the probabilities of the values of the other node. A positive qualitative influence of node $A$ on its successor $B$ expresses that observing higher values for $A$ makes higher values for $B$ more likely, regardless of any other direct influences on $B$; the influence is denoted $S_G^+(A, B)$, where '+' is the influence's *sign*. A negative qualitative influence, denoted

$S_G^-$, and a zero qualitative influence, denoted $S_G^0$, are defined analogously. If the influence of node $A$ on node $B$ is not monotonic or unknown, we say that it is *ambiguous*, denoted $S_G^?(A, B)$.

The set of influences of a qualitative probabilistic network exhibits various properties [Wellman, 1990]. The property of *symmetry* states that, if the network includes the influence $S_G^\delta(A, B)$, then it also includes $S_G^\delta(B, A)$, $\delta \in \{+, -, 0, ?\}$. The property of *transitivity* asserts that qualitative influences along a simple chain that specifies at most one incoming arc for each node, combine into a single influence with the $\otimes$-operator from Table 1. The property of *composition* asserts that multiple influences between two nodes along parallel chains combine into a single influence with the $\oplus$-operator.

Table 1: The $\otimes$- and $\oplus$-operators.

| $\otimes$ | + | − | 0 | ? |   | $\oplus$ | + | − | 0 | ? |
|-----------|---|---|---|---|---|----------|---|---|---|---|
| +         | + | − | 0 | ? |   | +        | + | ? | + | ? |
| −         | − | + | 0 | ? |   | −        | ? | − | − | ? |
| 0         | 0 | 0 | 0 | 0 |   | 0        | + | − | 0 | ? |
| ?         | ? | ? | 0 | ? |   | ?        | ? | ? | ? | ? |

In addition to influences, a qualitative probabilistic network includes *synergies* that express how the value of one node influences the probabilities of the values of another node in view of a value for a third node [Druzdzel & Henrion, 1993b]. A negative *product synergy* of node $A$ on node $B$ (and vice versa) given the value $c$ for their common successor $C$, denoted $X_G^-(\{A, B\}, c)$, expresses that, given $c$, higher values for $A$ render higher values for $B$ less likely. Positive, zero, and ambiguous product synergies are defined analogously. A product synergy induces a qualitative influence between the predecessors of a node upon observation of that node; the induced influence is coined an *intercausal influence*. In this paper, we assume that induced intercausal influences are added to a qualitative probabilistic network's graph as undirected edges.

**procedure** PropagateSign(*from*,*to*,*message*):

*sign[to]* ← *sign[to]* $\oplus$ *message*;
**for** each (induced) neighbour $V_i$ of *to*
**do** *linksign* ← sign of (induced) influence
            between *to* and $V_i$;
    *message* ← *sign[to]* $\otimes$ *linksign*;
    **if** $V_i \neq$ *from* **and** $V_i \notin$ *Observed*
        **and** *sign*[$V_i$] $\neq$ *sign*[$V_i$] $\oplus$ *message*
    **then** PropagateSign(*to*,$V_i$,*message*)

Figure 1: The Sign-propagation Algorithm.

For probabilistic inference with a qualitative probabilistic network, an elegant algorithm is available from M.J. Druzdzel and M. Henrion (1993a); this algorithm is summarised in pseudocode in Figure 1. The basic idea of the algorithm is to trace the effect of observing a node's value on the other nodes in a network by message-passing between neighbouring nodes. For each node, a *node sign* is determined, indicating the direction of change in the node's probability distribution occasioned by the new observation given all previously observed node values. Initially, all node signs equal '0'. For the newly observed node, an appropriate sign is entered, that is, either a '+' for the observed value *true* or a '−' for the value *false*, by calling PropagateSign(*observed node, observed node, sign*). Each node receiving a message updates its sign and subsequently sends a message to each neighbour that is not d-separated from the observed node and to every node on which it exerts an induced intercausal influence. The sign of this message is the ⊗-product of the node's (new) sign and the sign of the influence it traverses. This process is repeated throughout the network, building on the properties of symmetry, transitivity, and composition of influences. The process repeatedly visits each node that needs a change of sign. Since a node can change sign at most twice, once from 0 to + or −, and then only to ?, each node is visited at most twice. The process is therefore guaranteed to halt.

## 3 OUTLINE OF THE ALGORITHM

If a qualitative probabilistic network models trade-offs, it will typically yield ambiguous results upon inference with the sign-propagation algorithm. From Table 1, we have that whenever two conflicting influences on a node are combined with the ⊕-operator, an ambiguous sign will result. Once an ambiguous sign is introduced, it will spread throughout most of the network and an ambiguous sign is likely to result for the node of interest. By zooming in on the part of the network where the actual trade-offs reside and identifying the information that would serve to resolve these trade-offs, a more insightful result can be constructed. We illustrate the basic idea of our algorithm to this end.

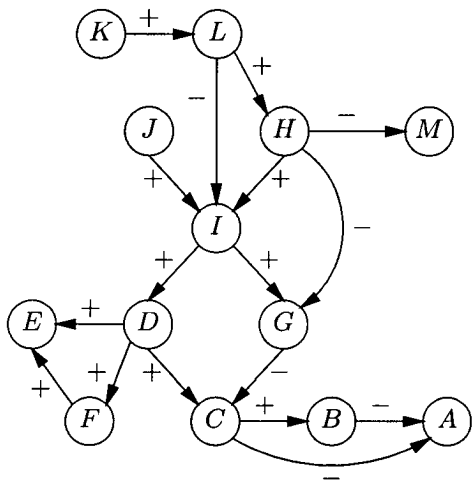

Figure 2: The Example Qualitative Probabilistic Network.

As our running example, we consider the qualitative probabilistic network from Figure 2. Suppose that the value *true*

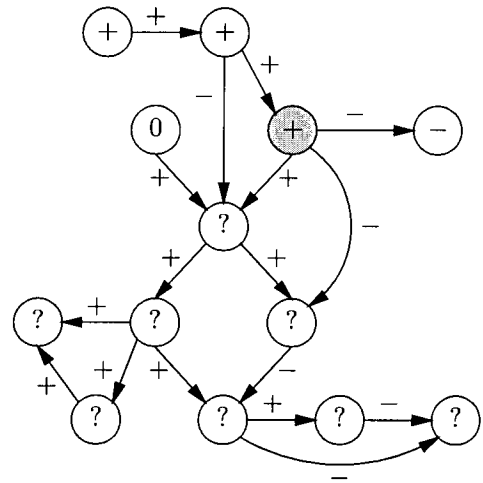

Figure 3: The Result of Propagating '+' for Node $H$.

has been observed for the node $H$ and that we are interested in its influence on the probability distribution of node $A$. Tracing the influence of the node sign '+' for node $H$, indicating its observed value, on every node's distribution by means of the sign-propagation algorithm, results in the node signs shown in Figure 3. These signs reveal that at least one trade-off must reside along the reasoning chains between the observed node $H$ and the node of interest $A$. These chains together constitute the part of the network that is relevant for addressing the trade-offs that have given rise to the ambiguous result for node $A$; we term this part the *relevant network*. For the example, the relevant network is shown in Figure 4 below the dashed line. Our algorithm now isolates this relevant network for further investigation. To this end, it deletes from the network all nodes and arcs that are connected to, but no part of the reasoning chains from $H$ to $A$.

A relevant network for addressing trade-offs typically includes many nodes with ambiguous node signs. Often, however, only a small number of these nodes are actually involved in the trade-offs that have given rise to the ambiguous result for the node of interest. Figures 3 and 4, for example, reveal that, while the nodes $A$, $B$, and $C$ have ambiguous node signs, the influences between them are not conflicting. In fact, *every possible* unambiguous node sign $sign[C]$ for node $C$ would result in the unambiguous sign $sign[C] \otimes ((+ \otimes -) \oplus -) = sign[C] \otimes -$ for node $A$. For addressing the trade-offs involved, therefore, the part of the relevant network between node $C$ and node $A$ can be disregarded. Node $C$ in fact separates the part of the relevant network that contains trade-offs from the part that does not. We call node $C$ the *pivot node* for the node of interest.

In general, the pivot node in a relevant network is a node with an ambiguous sign for which every possible unambiguous sign would uniquely determine an unambiguous sign for the node of interest; in addition, no other node having this property resides on an unblocked chain from the

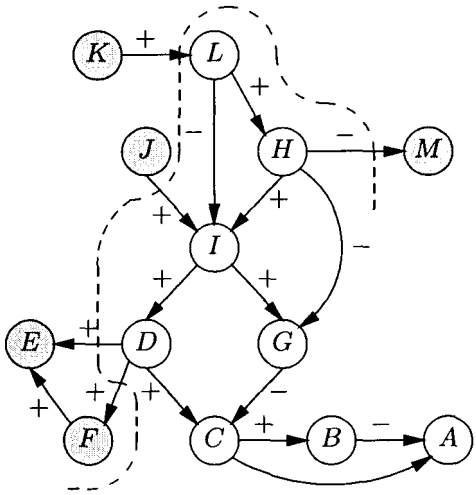

Figure 4: The Relevant Network, below the Dashed Line.

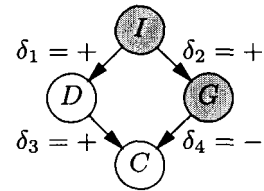

$$\delta_1 = + \qquad \delta_2 = +$$

$$\delta_3 = + \qquad \delta_4 = -$$

Figure 5: The Construction of a Sign for Node $C$.

observed node to the pivot node, that is, the pivot node is the node with this property "closest" to the observed node. Note that every network includes such a node. Our algorithm now selects from the relevant network the pivot node for the node of interest.

From the definition of pivot node, it can be shown that there must be two or more different reasoning chains in the relevant network from the observed node to the pivot node; the net influences along these reasoning chains, moreover, must be conflicting or ambiguous. To resolve the ambiguity at the pivot node, the relative strengths of the various influences as well as the signs of some of the nodes involved need be known. From Figures 3 and 4, for example, we have that node $I$ lies at the basis of the ambiguous sign for the pivot node $C$. Note that it receives an ambiguous node sign itself as a result of two conflicting (non-ambiguous) influences. An unambiguous node sign for node $I$ would not suffice to fix an unambiguous sign for node $C$. Even knowledge of the relative strengths of the two conflicting influences from node $I$ on the pivot node would not suffice for this purpose, however: a positive node sign for node $I$, for example, would still cause node $G$, residing on one of the reasoning chains from $I$ to $C$, to receive an ambiguous node sign, which in turn gives rise to an ambiguous influence on $C$. Node $G$ therefore also lies at the basis of the ambiguity at the pivot node. Now, every combination of unambiguous node signs for the nodes $G$ and $I$ would render the separate influences on the pivot node unambiguous. Knowledge of the relative strengths of these influences would suffice to determine an unambiguous sign for the pivot node. We call a minimal set of nodes having this property the *resolution frontier* for the pivot node.

In terms of signs for the nodes from the resolution frontier, our algorithm now constructs a (conditional) sign for the pivot node by comparing the relative strengths of the various influences exerted on it upon inference. In the example network, the nodes from the resolution frontier exert two separate influences on the pivot node $C$: the influence from node $I$ via node $D$ on $C$ and the influence from $G$ on $C$. For the sign $\delta$ of the influence of node $I$ via node $D$ on $C$ and for the sign $\delta'$ of the influence of $G$ on $C$, we find that

$$\begin{aligned} \delta &= sign[I] \otimes \delta_1 \otimes \delta_3 & \delta' &= sign[G] \otimes \delta_4 \\ &= sign[I] \otimes + & &= sign[G] \otimes - \end{aligned}$$

where $\delta_i$, $i = 1, 3, 4$, are as in Figure 5. For the node sign $sign[C]$ of the pivot node, the algorithm now constructs the following result:

if $|\delta| \geq |\delta'|$, then $sign[C] = \delta$, else $sign[C] = \delta'$;

where $|\delta|$ denotes the strength of the sign $\delta$. So, if the two influences on node $C$ have opposite signs, then their relative strengths will determine the sign for node $C$. The sign of the node of interest $A$ then follows directly from the node sign of $C$.

# 4   SPLITTING UP AND CONSTRUCTING SIGNS

In this section we detail some of the issues involved in our algorithm for pivotal pruning of trade-offs. In doing so, we assume that a qualitative probabilistic network does not include any ambiguous influences, that is, ambiguous node signs upon inference result from unresolved trade-offs. We further assume that observations are entered into the network one at a time. We also assume that sign propagation resulted in an ambiguous sign for the network's node of interest. For ease of reference, Figure 6 summarises the zoom algorithm in pseudocode.

**procedure** PivotalPruning($Q$):

$Q_{rel} \leftarrow$ ComputeRelevantNetwork($Q$);
$pivot \leftarrow$ ComputePivot($Q_{rel}$);
ConstructResults($Q_{rel}$,$pivot$)

Figure 6: The Basic Algorithm.

In detailing the algorithm, we focus attention on identifying the relevant part of a qualitative probabilistic network along with its pivot node and on constructing from these an informative result for the node of interest.

## 4.1 IDENTIFYING THE RELEVANT NETWORK

Our algorithm identifies from a qualitative probabilistic network the relevant part for addressing the trade-offs that have resulted in an ambiguous sign for the node of interest. We begin by formally defining the concept of relevant network.

**Definition 1** *Let $Q = (G, \Delta)$ be a qualitative probabilistic network as defined in Section 2. Let $O$ be the set of previously observed nodes in $Q$, let $E$ be the node for which new evidence has become available, and let $I$ be the network's node of interest. The* relevant network *for $E$ and $I$ given $O$ is the qualitative probabilistic network $Q_{rel} = (G', \Delta')$ such that*

- $V(G')$ *consists of all nodes that occur on a chain from $E$ to $I$ that is not blocked by $O$;*

- $A(G') = (V(G') \times V(G')) \cap A(G)$; *and*

- $\Delta'$ *consists of all qualitative influences and synergies from $\Delta$ that involve nodes from $G'$ only.*

The concept of relevance has been introduced before, most notably for quantitative probabilistic networks (see for example [Druzdzel & Suermondt, 1994, Shachter, 1998]). In fact, for quantitative and qualitative probabilistic networks various different concepts of relevance have been distinguished. For a node of interest $I$, previously observed nodes $O$, and a newly observed node $E$, we say that a node $N$ is

- *structurally relevant* to $I$, if $N$ is not d-separated from $I$ given $O \cup \{E\}$;

- *computationally relevant* to $I$, if the (conditional) probabilities for $N$ are required for computing the posterior probability distribution for $I$ given the observations for $O \cup \{E\}$; and

- *dynamically relevant* to $I$ and $E$, if $N$ partakes in the impact of $E$ on $I$ in the presence of the observations for $O$.

In our example qualitative network, node $D$ is structurally relevant, computationally relevant, and dynamically relevant to the node of interest $A$. Node $E$ is structurally relevant to node $A$ yet neither computationally nor dynamically relevant. Node $J$ is structurally irrelevant to the observed node $H$, as is also evidenced by its node sign '0' upon inference; it is both structurally and computationally relevant to the node of interest $A$, yet dynamically irrelevant. The newly observed node $H$ is d-separated from $A$ by its being observed. It therefore is not structurally relevant to $A$; it is computationally as well as dynamically relevant to $A$, however. Node $M$, to conclude, is neither structurally nor computationally nor dynamically relevant to the node of interest $A$.

The concept of dynamic relevance was introduced to denote the nodes constituting the reasoning chains between a newly observed node and a node of interest in a probabilistic network [Druzdzel & Suermondt, 1994]. The set of all nodes that are dynamically relevant to the node of interest $I$ and the newly observed node $E$, given the previously observed nodes $O$, can in fact be shown to induce the relevant network for $E$ and $I$ given $O$, as defined in Definition 1.

From a qualitative probabilistic network, the set of dynamically relevant nodes can be established by first determining all nodes that are computationally relevant to the node of interest $I$ and then removing the nodes that are not on any reasoning chain from the newly observed node $E$ to $I$. For computing the set of all computationally relevant nodes, the efficient *Bayes-Ball* algorithm is available from R.D. Shachter (1998). The algorithm takes for its input a probabilistic network, the set of all observed nodes $O \cup \{E\}$, and the node of interest $I$; it returns the sets of nodes that are computationally relevant, or *requisite*, to $I$. From the set of computationally relevant nodes, all nodes that are not on any reasoning chain from the newly observed node $E$ to the node of interest $I$ need be identified; these nodes are termed *nuisance nodes* for $E$ and $I$. An efficient algorithm is available for identifying these nodes [Lin & Druzdzel, 1997]. The algorithm takes for its input a computationally relevant network, the set of previously observed nodes $O$, the newly observed node $E$, and the node of interest $I$; it returns the set of nuisance nodes for $E$ and $I$. The algorithm for computing the relevant part of a qualitative probabilistic network is summarised in pseudocode in Figure 7.

---

**function** ComputeRelevantNetwork($Q$): $Q_{rel}$

*requisites* ← BayesBall($G, O \cup \{E\}, I$);
$V(G) \leftarrow (V(G) \setminus requisites) \cup \{E\}$;
$A(G) \leftarrow (V(G) \times V(G)) \cap A(G)$;
*nuisances* ← ComputeNuisanceNodes($G$);
$V(G) \leftarrow V(G) \setminus nuisances$;
$A(G) \leftarrow (V(G) \times V(G)) \cap A(G)$;
$\Delta \leftarrow \{$all influences and synergies from $\Delta$ in $G\}$;
**return** $Q_{rel} = (G, \Delta)$

---

Figure 7: The Algorithm for Computing the Relevant Network.

## 4.2 IDENTIFYING THE PIVOT NODE

After establishing the relevant part of a qualitative probabilistic network for addressing the trade-offs present, our algorithm identifies the pivot node. The pivot node serves to separate the part of the relevant network that contains the trade-offs that have given rise to the ambiguous sign for the node of interest, from the part that does not con-tain these trade-offs. The pivot node will allow for further focusing. We recall that the pivot node is a node with an ambiguous node sign, for which every possible unambiguous sign would uniquely determine an unambiguous sign for the node of interest. We define the concept of pivot node more formally.

**Definition 2** *Let* $Q = (G, \Delta)$ *be a relevant qualitative probabilistic network; let* $O$ *be the set of previously observed nodes, let* $E$ *be the newly observed node, and let* $I$ *be the network's node of interest, as before. The* pivot node *for* $I$ *and* $E$ *is a node* $P \in V(G)$ *such that*

- $S_G^\delta(E, P) \in \Delta$ *with* $\delta =$ '?';

- $S_G^{\delta'}(P, I) \in \Delta$ *with* $\delta' \neq$ '?'; *and*

- *there does not exist a node* $P'$ *with the above properties that resides on a chain from* $E$ *to* $P$ *that is not blocked by* $O$.

The pivot node in a relevant qualitative probabilistic network has various convenient properties. Before discussing these properties, we briefly review the concept of an *articulation node* from graph theory. In a digraph, an articulation node is a node that upon removal along with its incident arcs, makes the digraph fall apart into various separate components. In the digraph of our example network, as shown in Figure 2, the articulation nodes are the nodes $C, D, H, I$, and $L$; for the relevant network, depicted in Figure 4, node $C$ is the only articulation node, however. Articulation nodes are identified using a depth-first search algorithm; for details, we refer the reader to [Cormen *et al.*, 1990]. Theorems 1 and 2 now state important properties of a pivot node that allow for its identification.

**Theorem 1** *Let* $Q = (G, \Delta)$ *be a relevant qualitative probabilistic network; let* $E$ *be the newly observed node and let* $I$ *be the node of interest. The pivot node for* $I$ *and* $E$ *is either the node of interest* $I$ *or an articulation node in* $G$.

**Proof** (sketch). By definition we have that every possible unambiguous sign for the pivot node determines an unambiguous sign for the node of interest $I$. It will be evident that node $I$ itself satisfies this property. Either the node of interest $I$ or another node on an unblocked chain from $E$ to $I$, therefore, is the pivot node. Now, suppose that node $I$ is not the pivot node. As a sign for the pivot node uniquely determines the sign for $I$, we conclude that all influences exerted upon $I$ must traverse the pivot node. Every unblocked chain from $E$ to $I$, therefore, must include the pivot node. As a consequence, removing the pivot node along with its incident arcs from the relevant network will cause the network to fall apart into separate components. We conclude that the pivot node is an articulation node. $\square$

**Theorem 2** *Let* $Q = (G, \Delta)$ *be a relevant qualitative probabilistic network; let* $E$ *and* $I$ *be as before. The pivot node for* $I$ *and* $E$ *is unique.*

**Proof** (sketch). From Definition 1 we have that the relevant network consists of only nodes that reside on an unblocked chain from the newly observed node $E$ to the node of interest $I$. From the definition of articulation node, we further have that every such chain must include all articulation nodes in the relevant network. In fact, every reasoning chain from $E$ to $I$ visits the articulation nodes in the same order. From Definition 2 we have that no two pivot nodes can reside on the same unblocked chain to the node of interest. We conclude that the pivot node is unique. $\square$

From the proof of Theorem 2 we have that the articulation nodes in a relevant network allow a total ordering. We number the articulation nodes, together with the node of interest $I$, from 1, for the node closest to the newly observed node, to $m$, for the node of interest. The pivot node now is the node with the lowest ordering number for which an unambiguous sign would uniquely determine an unambiguous sign for the node of interest. To identify the pivot node, our algorithm starts with investigating the articulation node closest to the node of interest; this node is numbered $m - 1$. The algorithm investigates whether an unambiguous sign for this candidate pivot node would result in an unambiguous sign for the node of interest upon sign propagation. By propagating a '+' from the candidate pivot node to the node of interest $I$, the node sign resulting for $I$ is the sign of the net influence of the candidate pivot node on $I$. If this sign is ambiguous, then the node of interest itself is the pivot node. Otherwise, the algorithm proceeds by investigating the articulation node numbered $m - 2$, and so on. The algorithm is summarised in pseudocode in Figure 8.

```
function ComputePivot(Q): pivot
candidates ← {I} ∪ FindArticulationNodes(G);
order the nodes from candidates from 1 to m;
return FindPivot(m − 1);

function FindPivot(i): pivot
PropagateSign(node i,node i,'+')
if sign[node i + 1] = '?'
then return node i + 1;
else FindPivot(i − 1)
```

Figure 8: The Algorithm for Computing the Pivot Node.

### 4.3 CONSTRUCTING RESULTS

From its definition, we have that there must be two or more different reasoning chains in the relevant network from the newly observed node to the pivot node; the net influences along these reasoning chains are conflicting or ambiguous. Our algorithm focuses on the ambiguity at the pivot node and identifies the information that would serve to resolve it. For this purpose, the algorithm zooms in on the part of the relevant network between the newly observed node and the pivot node; we call this part the *pruned relevant network*. Note that the pruned relevant network is readily computed by exploiting the property that the pivot node is an articulation node. From the pruned relevant network, the algorithm first selects the so-called *candidate resolvers*.

**Definition 3** *Let* $Q = (G, \Delta)$ *be a relevant qualitative probabilistic network; let* $E$ *be the newly observed node and let* $I$ *be the network's node of interest. Let* $P$ *be the pivot node for* $I$ *and* $E$. *Now, let* $Q_{pru} = (G', \Delta')$ *be the pruned relevant network for* $P$. *A candidate resolver for* $P$ *is a node* $R_i \in V(G')$, $R_i \neq P$, *such that*

- $R_i = E$, *or*
- $sign[R_i] =$ '?' *and in-degree* $[R_i] \geq 2$.

The candidate resolvers for the pivot node are easily identified from the pruned relevant network.

From among the candidate resolvers in the pruned relevant network, our algorithm now constructs the resolution frontier. We recall that the resolution frontier is a minimal set of nodes for which unambiguous node signs would uniquely determine the signs of the separate influences on the pivot node.

**Definition 4** *Let* $Q = (G, \Delta)$ *be a pruned relevant qualitative probabilistic network; let* $E$ *and* $I$ *be as before. Let* $P$ *be the pivot node for* $I$ *and* $E$, *and let* $R$ *be the set of candidate resolvers for* $P$, *as defined in Definition 3. The resolution frontier* $F$ *for* $P$ *is the maximal subset of* $R$ *such that for each candidate resolver* $R_i \in F$ *there exists at least one unblocked chain from* $E$ *via* $R_i$ *to* $P$ *such that no node* $R_j \in R$ *resides on the subchain from* $R_i$ *to* $P$.

The resolution frontier can be constructed by recursively traversing the various reasoning chains from the pivot node back to the observed node $E$ and checking whether the nodes visited are candidate resolvers.

Once the resolution frontier has been identified from the pruned relevant network, the algorithm constructs a (conditional) sign for the pivot node in terms of signs for the nodes from the frontier. Let $F$ be the resolution frontier for the pivot node $P$. For each resolver $R_i \in F$, let $s_j^i$, $j \geq 1$, denote the signs of the various different reasoning chains from $R_i$ to the pivot node. For each combination of node signs $sign[R_i]$, $R_i \in F$, the sign of the pivot node is computed to be

$$\text{if } \left| \oplus_{(sign[R_i] \otimes s_j^i) = +} \left( sign[R_i] \otimes s_j^i \right) \right| \geq$$

$$\left| \oplus_{(sign[R_i] \otimes s_j^i) = -} \left( sign[R_i] \otimes s_j^i \right) \right| \quad (1)$$

$$\text{then } sign[P] = +, \text{ else } sign[P] = -$$

where $|\delta|$ once again is used to denote the strength of the sign $\delta$. We would like to note that as, in general, the resolution frontier includes a small number of nodes, the number of signs to be computed for the pivot node is limited. In addition, we note that the process of constructing informative results can be repeated recursively for the nodes in the pivot node's resolution frontier, until the newly observed node is reached. The basic algorithm is summarised in pseudocode in Figure 9.

**procedure** ConstructResults($Q$,*pivot*):

$Q_{pru} \leftarrow$ ComputePrunedNetwork($Q$,*pivot*);
*candidates* $\leftarrow$ ComputeCandidates($Q_{pru}$,*pivot*);
output ComputeResults($Q_{pru}$,*pivot*,*candidates*)

**function** ComputeResults($Q_{pru}$,*pivot*,*candidates*):

*frontier* $\leftarrow$ ComputeFrontier(*pivot*,$\varnothing$,*candidates*);
**for all** $R_i \in$ *frontier*
**do** determine $s_j^i$, $j \geq 1$;
**for all** $R_i \in$ *frontier* and $sign[R_i] = +, -$
**do return** inequality (1);

**function** ComputeFrontier(*pivot*,*frontier*,
*candidates*): *frontier*

**for all** $V_i$ such that $(V_i, pivot)$ or $(pivot, V_i)$
on a reasoning chain from $E$
**do if** $V_i \in$ *candidates*
**then** *frontier* $\leftarrow$ *frontier* $\cup \{V_i\}$
**else** ComputeFrontier($V_i$,*frontier*,*candidates*)

Figure 9: The Algorithm for Constructing Results.

To conclude, we would like to note that for computing informative results for a relevant network's pivot node, the pruned network can be even further restricted. To this end, a so-called *boundary node* can be identified for the newly observed node. The boundary node is the articulation node closest to the node of interest that has an unambiguous node sign after propagation of the observation entered. Constructing results can then focus on the part of the relevant network between the pivot node and the boundary node. Moreover, if the thus pruned network includes many articulation nodes, it may very well be that trade-offs exist between the articulation nodes numbered $k - 1$ and $k$, but not between $k$ and $k + 1$. Distinguishing between these components is straightforward and allows for further focusing on the actual trade-offs involved in inference.

## 5 CONCLUSIONS

We have presented a new algorithm for dealing with trade-offs in qualitative probabilistic networks. Rather than resolve trade-offs by providing for a finer level of representation detail, our algorithm identifies from a qualitative probabilistic network the information that would serve to resolve the trade-offs present. For this purpose, the algorithm zooms in on the part of the network where the actual trade-offs reside and identifies the pivot node for the node of in-terest. The sign of the pivot node uniquely determines the sign of the node of interest. For the pivot node, a more informative result than ambiguity is constructed in terms of values for the node's resolvers and the relative strengths of the influences upon it. This process of constructing informative results can be repeated recursively for the pivot node's resolvers.

As we have already mentioned in our introduction, we believe that qualitative probabilistic networks can play an important role in the construction of quantitative networks for real-life application domains, as well as for explanation of their reasoning processes. For the purpose of explanation, qualitative probabilistic networks have been proposed before. The concept of pivot node for zooming in on trade-offs and constructing insightful results for a network's node of interest is a very powerful concept to enable explanation of complex reasoning processes in quantitative probabilistic networks.

### Acknowledgments

This work was partially supported by the EPSRC under grant GR/L84117 and a Ph.D. studentship.